\documentclass{sig-alternate-05-2015}

\usepackage{graphicx}
\usepackage{booktabs}
\usepackage{amsmath}
\usepackage{flushend}
\def\y{{\bf y}}
\def\R{{\mathbb{R}}}
\def\x{\mathbf{x}}
\def\sharedaffiliation{%
\end{tabular}
\begin{tabular}{c}}

\begin{document}

\title{Improved Recurrent Neural Networks for Session-based Recommendations}

\numberofauthors{3}
\author{
  \alignauthor Yong Kiam Tan\\
  \email{tanyongkiam@gmail.com}
  \alignauthor Xinxing Xu\\
  \email{\mbox{xuxinx@ihpc.a-star.edu.sg}}
  \alignauthor Yong Liu\\
  \email{\mbox{liuyong@ihpc.a-star.edu.sg}}
  \sharedaffiliation
  \affaddr{Institute of High Performance Computing, A*STAR}  \\
  \affaddr{1 Fusionopolis Way, Singapore 138632}   \\
}

\CopyrightYear{2016}
\setcopyright{licensedothergov}
\conferenceinfo{DLRS '16,}{September 15 2016, Boston, MA, USA}
\isbn{978-1-4503-4795-2/16/09}\acmPrice{\$15.00}
\doi{http://dx.doi.org/10.1145/2988450.2988452}

\maketitle
\begin{abstract}
Recurrent neural networks (RNNs) were recently proposed for the session-based recommendation task. The models showed promising improvements over traditional recommendation approaches.
In this work, we further study RNN-based models for session-based recommendations.
We propose the application of two techniques to improve model performance, namely, data augmentation, and a method to account for shifts in the input data distribution.
We also empirically study the use of generalised distillation, and a novel alternative model that directly predicts item embeddings.
Experiments on the RecSys Challenge 2015 dataset demonstrate relative improvements of 12.8\% and 14.8\% over previously reported results on the Recall@20 and Mean Reciprocal Rank@20 metrics respectively.
\end{abstract}

%
%

\ccsdesc[500]{Computing methodologies~Supervised learning}
\ccsdesc[500]{Computing methodologies~Neural networks}
\ccsdesc[500]{Information systems~Recommender systems}

%
%

\printccsdesc

\keywords{Recurrent neural networks; Recommender systems; Session-based recommendations}

\section{Introduction}
Users of e-commerce websites are often inundated by the huge number of items available for sale. Recommender systems can be used to enhance user experience by making personalized and useful recommendations for each user. For example, the system could automatically display items of interest, or suggest new discounts relevant to each user. In order to personalize recommendations, traditional recommender systems often need to build up a user profile. Collaborative filtering approaches~\cite{MFTRS,Koren08,RBMCF} can define user-user similarity based on their history of purchases, or they could rely on matrix factorization to build latent factor vectors for each user. Crucially, these approaches require the user to be identified when making recommendations. This may not always be possible: new users to the site will not have any profile, or users may not be logged in, or they may have deleted their tracking information. This leads to the problem of cold-start for recommendation methods that require user history.

An alternative to relying on historical data is to make session-based recommendations~\cite{ephemeral}. In this setting, the recommender system makes recommendations based only on the behaviour of users in the current browsing session. This avoids the aforementioned cold-start issue but we must ensure that the system remains accurate and responsive (i.e.~the predictions do not take too long to make). Recurrent Neural Networks (RNNs) were recently proposed in~\cite{ICLR16} for the session-based recommendation task. The authors showed significant improvements over traditional session-based recommendation models using an RNN. The proposed model utilizes session-parallel mini-batch training, and also employs ranking-based loss functions for learning the model.

In this work, we further study the application of RNNs for session-based recommendations. In particular, we examine and adapt various techniques from the literature for this task. These include:
\begin{itemize}
\item Data augmentation via sequence preprocessing and embedding dropout to enhance training and reduce over-fitting.
\item Model pre-training to account for temporal shifts in the data distribution.
\item Distillation using privileged information to learn from small datasets.
\end{itemize}
Additionally, we propose a novel alternative model that reduces the time and space requirements for predictions by predicting item embeddings directly. This makes RNNs more readily deployable in real-time settings.

Our proposed techniques were evaluated on the RecSys Challenge 2015 data set. The effectiveness of our data augmentation strategy is evidenced by relative model performance improvements of 12.8\% and 14.8\% over previously reported results on the Recall@20 and Mean Reciprocal Rank\\@20 (MRR@20) metrics respectively. We also showed that distillation could be successfully applied for performance gains on small datasets. Finally, our novel item embedding output approach significantly reduces the time and space requirements of the RNN model.

We start with a discussion of related work in Section~\ref{sec:Rework}. Then, we present the details of our improved RNN models in Section~\ref{sec:IRNN}, and our experiments on the models in Section~\ref{sec:experiments}.

\section{Related Work}\label{sec:Rework}

Matrix factorization and neighbourhood-based methods are widely utilized for recommender systems in the literature.
Matrix factorization methods~\cite{MFTRS,WeimerKLS07} are based on the sparse user-item interaction matrix, where the recommendation problem is formulated as a matrix completion task. After decomposing the matrix, each user and item is represented by a latent factor vector. The missing value of the user-item matrix can then be filled by multiplying the appropriate user and item vectors. Since this requires us to identify both the user and item vectors, matrix factorization methods are not directly suitable for session-based recommendations where the users are unknown. One way to solve this cold-start problem is to use pairwise preference regression~\cite{pairwise}. Neighbourhood based methods~\cite{SarwarKKR01,Koren08} utilize the similarities between item and user purchase history; they can be applied to session-based recommendations by comparing session similarity.

Deep learning has recently been applied very successfully in areas such as image recognition~\cite{ImageNet,DeepResi}, speech recognition~\cite{GravesMH13,DeepSpeech2} and natural language processing~\cite{RNNNLP}. Deep models can be trained to learn discriminative representmations from unstructured data such as images and speech signals. They have also been used for collaborative filtering~\cite{CollabDL,RBMCF}. In~\cite{ICLR16}, RNNs were proposed for session-based recommendations. The authors compared RNNs (with several customized ranking losses) to existing methods for session-based predictions and found that RNN-based models performed 20\% to 30\% better than the baselines. Our work is closely related, and we study extensions to their RNN models. In~\cite{clickseqprediction}, the authors also use RNNs for click sequence prediction; they consider historical user behaviours as well as hand engineered features for each user and item. In this work, we rely entirely on automatically learned feature representations.

Many approaches have been proposed to improve the predictive performance of trained deep neural networks. Popular approaches include data augmentation~\cite{ImageNet}, dropout~\cite{dropout,YarinArxiv}, batch normalization~\cite{BNorm} and residual connections~\cite{DeepResi}. We seek to apply some of these methods to enhance the training of our recommendation RNNs.

The learning using privileged information (LUPI) framework~\cite{SVM+,SVM2+,THash} was proposed to utilize the additional feature representations that are only available during training but not during testing. When there is a limited amount of training data, the use of such information has been found to be helpful~\cite{SVM+}. In the generalized distillation approach~\cite{distillation}, a student model learns from soft labels provided by a teacher model. If we train the teacher model on the privileged dataset, then this approach can be applied to LUPI. In this work, we propose the use of this framework for the click sequence prediction by using the future portion of each click sequence as a form of privileged information.

\section{Proposed Approaches}\label{sec:IRNN}

In this section, we explain the use of RNNs for the session-based recommendation problem~(\ref{RNNModel}). This is followed by our proposed data augmentation methods~(\ref{preprocess}), our approach to handling temporal shifts~(\ref{temporal}), an explanation of the application of LUPI~(\ref{privileged}), and finally, an alternative model based on embeddings to trade model accuracy for speed and memory requirements~(\ref{embedding}).

\subsection{RNNs for session-based recommendations}\label{RNNModel}
The session-based recommendation problem can be formulated as a sequence-based prediction problem as follows. Let $[x_1,x_2,\ldots,x_{n-1},x_{n}]$ be a click session, where $x_i\in \R$ ($1 \leq i \leq n$) is the index of one clicked item out of a total number of $m$ items.

We seek a model $M$ such that for any given prefix click-sequence of the session, $\mathbf{x} = [x_1,x_2,\ldots,x_{r-1},x_{r}]$, $1 \leq r < n$, we get the output $\y = M(\mathbf{x})$, where $\y = [y_1,\ldots,y_m]' \in \R^{m}$. We view $\y$ as a ranking over all the next items that can occur in that session, where $y_i$ corresponds to the score of item $i$. Since we typically need to make more than one recommendation for the user to choose from, the top-$k$ items (as ranked by $\y$) are recommended.

In most of our models, we use a classification-based output, where $\y$ corresponds to a probability distribution over the items. Let $x_{r+1}$ be the next click of the click sequence $\mathbf{x}$; we can represent it with an $m$-dimensional 1-HOT encoded vector $V(x)\in \R^{m}$. The model can be tuned by minimizing a chosen loss function e.g.~the cross entropy loss, $L(M(\mathbf{x}),V(x_{r+1}))$. Other outputs are possible: the models in~\cite{ICLR16} output ranking scores for each item, and they are trained with ranking losses.

\begin{figure}
\centering
\includegraphics[trim = 280 50 280 50,width=0.42\linewidth,clip] {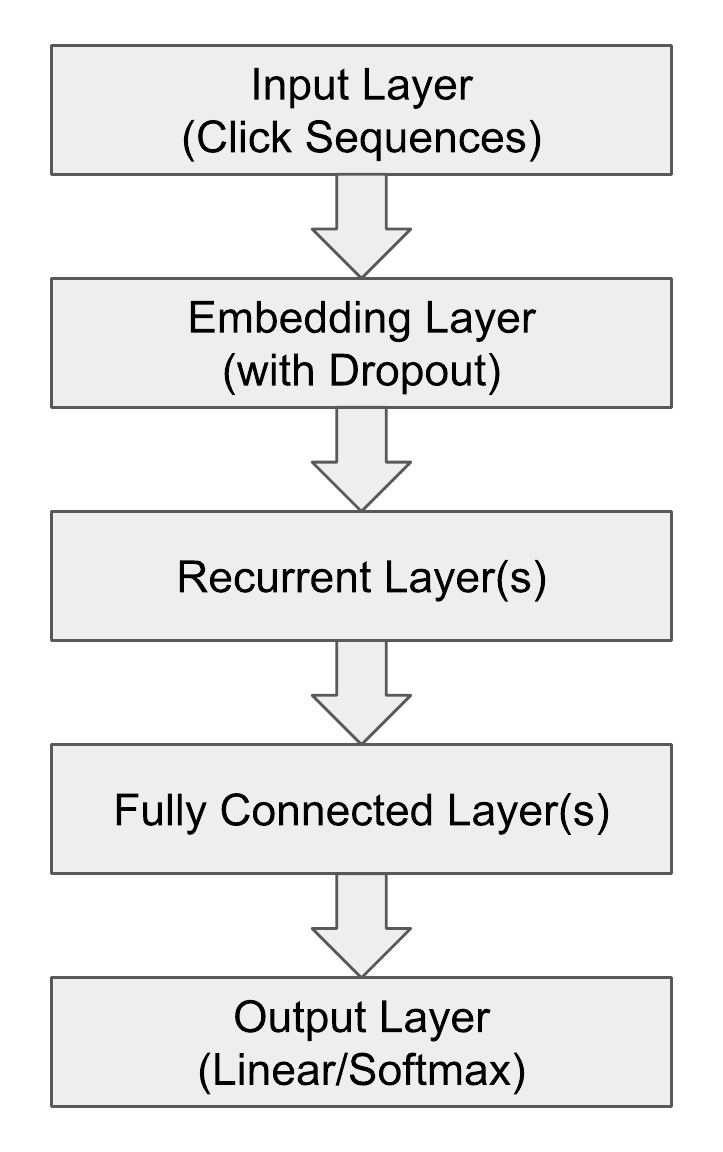}
\caption{Generic structure of the network used in our models. The output layer can either use a softmax or linear activation function.}
\label{genericmodel}
\end{figure}

\begin{figure}
\centering
\includegraphics[trim = 100 40 100 40,width=0.8\linewidth,clip] {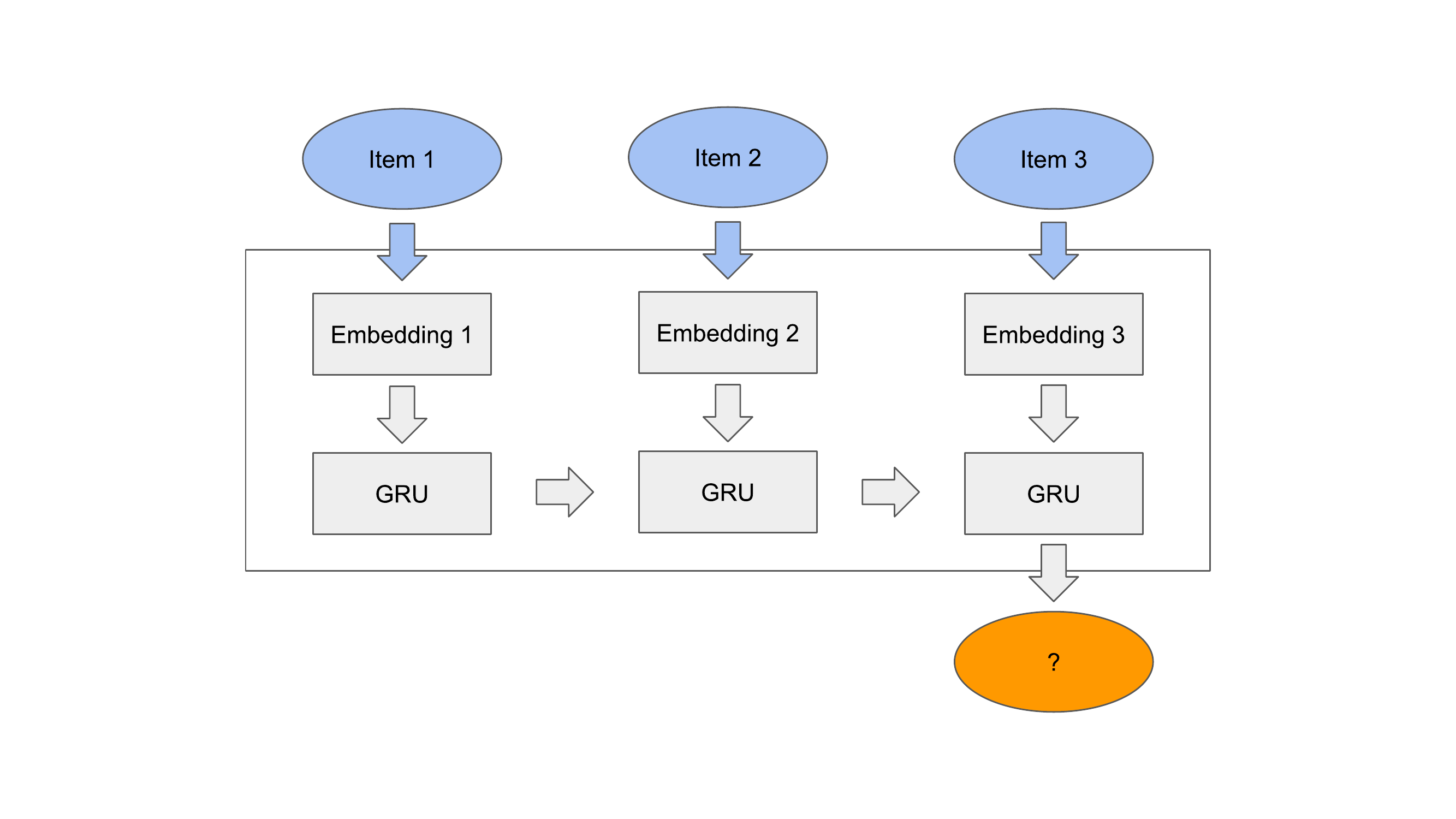}
\caption{Training procedure for a single sequence in our RNN. Gradients are backpropagated along the grey arrows. The input item sequence (in blue) and target output (in orange) are typically provided in mini-batches.}
\label{trainingprocedure}
\end{figure}

We follow the generic structure of the RNN model shown in Figure~\ref{genericmodel}. For the recurrent layers, we use the Gated Recurrent Unit (GRU)~\cite{GRU} as it was found in~\cite{ICLR16} that they outperformed the Long-term Short Memory (LSTM)~\cite{LSTM} units. However, we do not utilize the stateful RNN training procedure, where the models are trained in a session-parallel, sequence-to-sequence manner. Instead, our networks process each sequence $[x_1,x_2,\ldots,x_{r}]$ separately, and are trained to predict the next item, $x_{r+1}$, in that sequence. We also represent all our input using trainable embeddings. Our networks can be trained using standard mini-batch gradient descent on the cross-entropy loss via Backpropagation-Through-Time (BPTT) for a fixed number of time steps. This training procedure is visualized in Figure~\ref{trainingprocedure}.

\subsection{Data augmentation}
\label{preprocess}
Click sessions often vary in length: some users may take a long time before finding their desired item, while others find it with just a few clicks. One aim of the recommender system should be to provide accurate predictions regardless of the current session length. Data augmentation techniques have been widely used to enhance image-based models~\cite{ImageNet}. Here, we propose two methods to augment click sequences.

The first is an application of the sequence preprocessing method proposed in~\cite{AlexandreArxiv}. All prefixes of the original input sessions are treated as new training sequences. Given an input training session $[x_1,x_2,\ldots,x_{n}]$, we generate the sequences and corresponding labels $([x_1],V(x_2))$, $([x_1,x_2],V(x_3))$, \ldots, $([x_1,x_2,\ldots,x_{n-1}],V(x_n))$ for training.

Embedding dropout is a form of regularization applied to input sequences \cite{YarinArxiv}. Applying it to a click sequence is equivalent to a preprocessing step that randomly deletes clicks at random. Intuitively this makes our model less sensitive to noisy clicks, e.g.~where users may have accidentally clicked on items that are not of interest. Hence, it makes the model less likely to over-fit to specific noisy sequences. It can also be viewed as a form of data augmentation, where shorter, pruned sequences are being generated for model training.

\begin{figure}
\centering
\includegraphics[clip,trim = 150 0 150 0,width=0.9\linewidth]{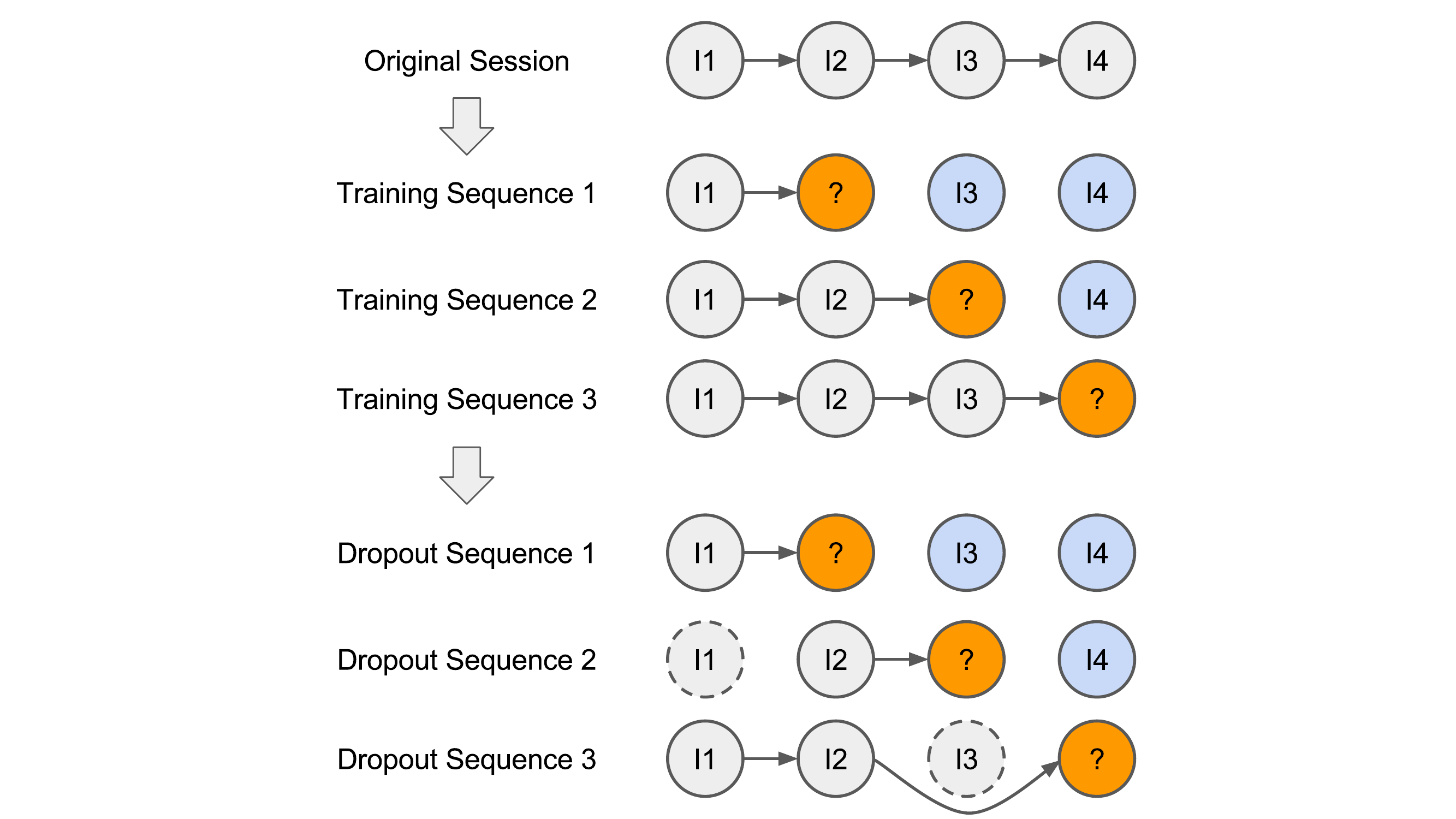}
\caption{An example application of our preprocessing step on a session with four clicks. The items in orange are output labels corresponding to their respective training sequences (in grey), and the items with a dotted outline are randomly dropped during training. The privileged information for each preprocessed sequence is coloured in blue, they are not used in the standard training procedure.}
\label{examplepreprocess}
\end{figure}

We apply both methods to all our models, and a graphical example is shown in Figure~\ref{examplepreprocess}. Note that different clicks are dropped in each sequence for every training epoch.

\subsection{Adapting to temporal changes}
\label{temporal}
A key assumption of many machine learning models is that the input is independent and identically distributed. This is not strictly true in the item recommendation setting since new products will only appear in sessions collected after that product is released, and user behaviour/preferences may also shift over time. Moreover, the purpose of the recommender system is to make prediction on new sequences, i.e.~those arising from recent user behaviours. Learning a recommendation model on the entire dataset may, therefore, lead to worse performance because the model ends up focusing on some out-of-date properties that are irrelevant to the latest sequences. One way to handle this is to define a temporal threshold, and discard click sequences that are older than the threshold when building the model. However, this reduces the amount of training data available for our models to learn from.

We propose a simple solution to get the best of both worlds via pre-training. We first train a model on the entire dataset. The trained model is then used to initialize a new model, which is only trained using only a more recent subset of the data, e.g.~the last month worth of data out of a year of click sequences. This allows the model to have the benefit of a good initialization using large amounts of data, and yet is focused on more recent click-sequences. In this way, it resembles the fine-tuning process used in training of image-based networks~\cite{Chatfield14}, where the models are typically initialized by pre-training on ImageNet (a large image classification dataset) before the weights are fine-tuned on a smaller image dataset in the desired domain.

\subsection{Use of privileged information}
\label{privileged}
The item sequence clicked by users \emph{after} an item may also contain information about that item (highlighted in Figure~\ref{examplepreprocess}). This information cannot be used for making predictions since we cannot view the future sequences when making recommendations. We can, however, utilize these future sequences as \emph{privileged information}~\cite{SVM+} in order to provide soft labels for regularizing and training our models. We use the generalized distillation framework \cite{LUPI} for this purpose.

Formally, given a sequence $[x_1,x_2,\ldots,x_{r}]$ with label $x_{r+1}$ from a session, we define the privileged sequence as $\mathbf{x^*} = [x_{n},x_{n-1},\ldots,x_{r+2}]$ where $n$ is the length of the original session before our preprocessing. The privileged sequence is simply the reversed, future sequence that occurs after the $r^{th}$ item. We can now train a teacher model on the privileged sequences $\mathbf{x^*}$, with the same label, $x_{r+1}$.

Next, we tune our student model $M(\mathbf{x})$ by minimizing a loss function\footnote{The original presentation also includes a temperature parameter $T$ which can be used to control the softness of the softmax labels.} of the form: $(1-\lambda) * L(M(\mathbf{x}),V(x_n)) + \lambda * L(M(\mathbf{x}),M^*(\mathbf{x^*}))$, where $\lambda\in[0,1]$ is a tradeoff parameter between the two sets of labels. This allows $M$ to learn from both the real labels, as well as the labels predicted by its teacher, $M^*$. This learning procedure is useful when the amount of training data available is small, which may be the case for a new, small scale website.

\subsection{Output embeddings for faster predictions}
\label{embedding}
An issue with the models we have described thus far is the size of the output layer. The output layer\footnote{Although we focus on the softmax activation function, this also applies for ranking-based outputs.} is typically fully connected to the previous hidden layer -- this means that the number of parameters to be tuned in these two layers alone is $H * N$ where $H$ is the number of nodes in the hidden layer and $N$ is the number of candidate items for prediction. Besides the memory requirements, this also makes prediction slower since the model has to perform an additional large matrix multiplication.

A similar problem has also been studied in natural language processing, where the output vocabulary can be huge. Typical approaches include the use of a hierarchical softmax layer \cite{NIPS2008}, and sampling only the most frequent items. The hierarchical softmax approach does not apply directly in our case, since we are required to make a top-$k$ prediction, rather than just a top-1 prediction.

We instead view item embeddings as a projection of the items from a 1-HOT encoded space of dimension $N$ onto a lower dimensional space. Using this point of view, we propose to train the model to predict the \emph{embedding} of the next item directly. The model is tuned using the cosine loss between the embedding of the true output and the predicted embedding. This approach is inspired by the distributed representations of words~\cite{word2vec}, where similar words have embeddings that are closer in cosine distance. We expect, similarly, that the items which a user is likely to click after a given sequence should be close in the item embedding space. Using this type of output reduces the number of parameters in the final layers to $H * D$, where $D$ is the dimensionality of the embedding. A drawback of this approach is that it requires a good quality embedding for each item. One way to obtain such an embedding is to extract and re-use the trained item embedding from the models described above.

\section{Experiments}\label{sec:experiments}

We evaluate our proposed extensions to the basic RNN model on the RecSys Challenge 2015 dataset. The dataset is split following \cite{ICLR16}, where sessions in the last day are placed in the test set, and everything else is placed in the training set. This yields 7966257 sessions in the training set, 15234 sessions in the test set, and 37483 candidate items for prediction. We have 23670981 training sequences after preprocessing the sessions. To better evaluate some of our models, e.g.~privileged information and pre-training, we sort the training sequences by time and report our results on models trained on more recent fractions ($\frac{1}{256},\frac{1}{64},\frac{1}{16},\frac{1}{4},\frac{1}{1}$) of the training sequences as well.

We also follow the evaluation procedure of \cite{ICLR16}; each session is input item-by-item to the model, and we calculate the model's ranking of the next item in the session. The evaluation metrics used were Recall@20 and Mean Reciprocal Rank (MRR)@20. These metrics are designed for the recommendation setting, as we usually want to make multiple recommendations for each user. For M1-M3, we take the top 20 most probable items directly from the softmax outputs. For M4, we compute the cosine distance of the model output against the embedding of items, and take the top 20 closest items\footnote{This can be efficiently computed on the GPU as an additional Theano expression.}. Finally, we also report the model size and batch prediction times for each model. These are important considerations if the model is going to be deployed in a real recommender system.


\begin{figure*}[t]
\centering
\includegraphics[trim = 40 70 40 70,width= 1\linewidth,clip] {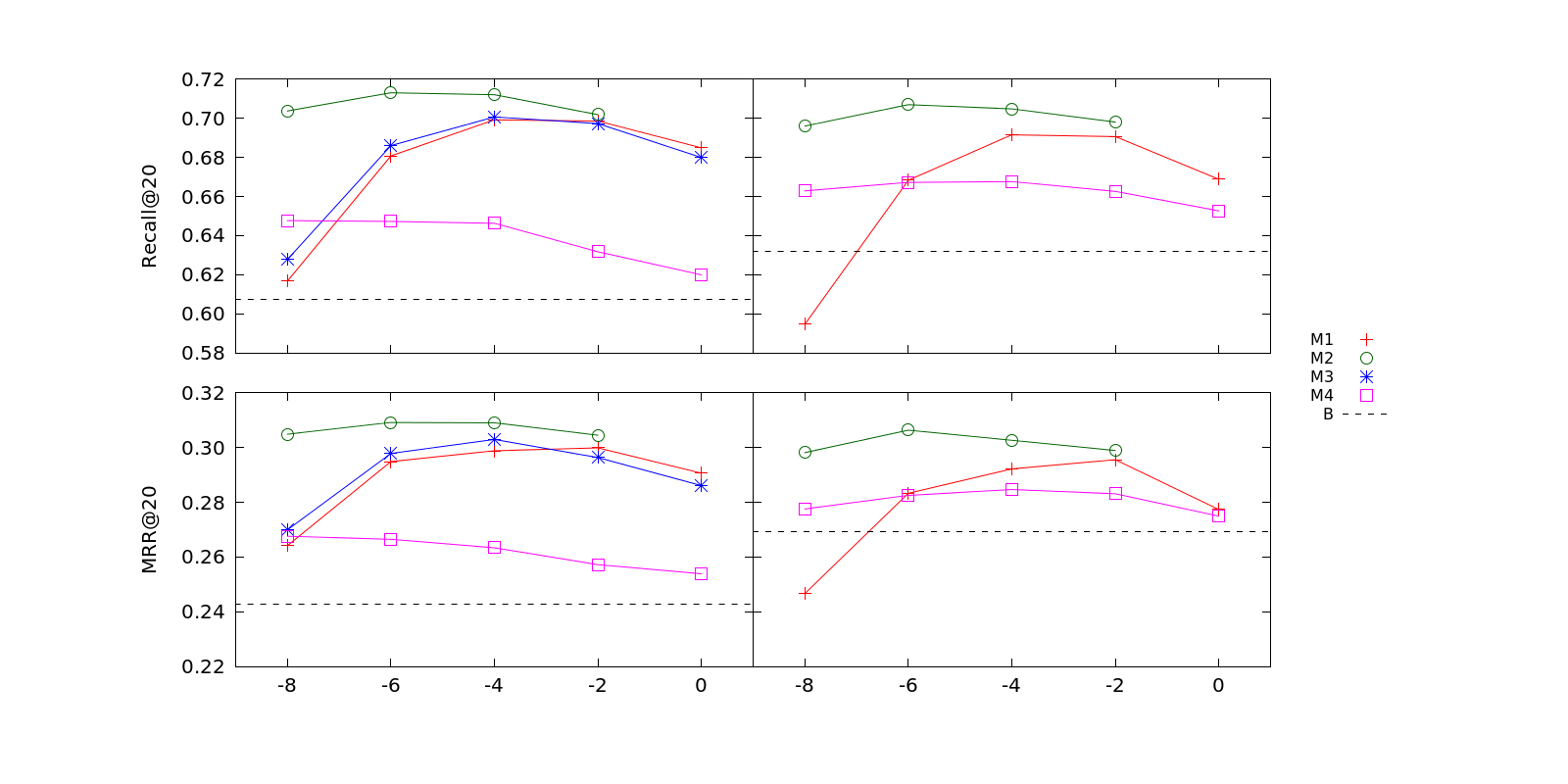}
\caption{Plots of both evaluation metrics on models with GRU size 100 (left) and 1000 (right). The x-axis is logarithmic in dataset fraction, the rightmost point corresponds to the full dataset. M2 does not apply to the full dataset, and results for M3 on the larger GRU size were omitted.}
\label{resplot}
\end{figure*}

\subsection{Experimental setup}
All our models used 50-dimensional embeddings for the items, with 25\% embedding dropout. Optimization was done using Adam \cite{Adam}, with mini-batch size fixed at 512. We truncated BPTT using a fixed window of 19 time-steps since 99\% of the original training sessions had lengths less than or equal to 19. Sequences shorter than 19 items were padded with zeros for simplicity, the RNN ignores these zeroes. The number of epochs was set by early stopping using 10\% of the training data as the validation set for each model. We used one recurrent (GRU) layer in all our models as we found that additional layers did not improve performance. The GRU was set at 100 and 1000 hidden units for each model. The models are defined and trained in Keras \cite{chollet2015keras} and Theano \cite{Theano} on a GeForce GTX Titan Black GPU. The specifics of each model (along with their labels) are as follows:

\begin{itemize}
\item[\textbf{M1}] The RNN model with softmax outputs, sequence preprocessing and embedding dropout. The recurrent layer is fully connected to the output layer.
\item[\textbf{M2}] M1, but additionally re-tuned on more recent fractions of the training dataset.
\item[\textbf{M3}] An M1 model trained on the privileged information (future sequences) available in each data fraction. This is used to provide soft labels for another M1 model with parameters $T=1$ and $\lambda=0.2$. We did not extensively tune these parameters.
\item[\textbf{M4}] The output of this model predicts an item embedding directly. We added a fully connected hidden layer between the recurrent and output layers as we found that this improved the model's stability. We used the embeddings trained on the full training dataset in M1 for these models.
\item[\textbf{B}] This refers to the best results reported in \cite{ICLR16}.
\end{itemize}

\subsection{Experimental results}

\begin{table}[t]
\centering
\begin{tabular}{@{}ccc@{}}
\toprule
\textbf{Model Type (GRU Size)}   & \textbf{Recall@20} & \textbf{MRR@20} \\ \midrule
S-POP (-) \cite{ICLR16}          & 0.2672               & 0.1775            \\
Item-KNN (-) \cite{ICLR16}       & 0.5065               & 0.2048            \\
TOP1 (1000) \cite{ICLR16}        & 0.6206               & 0.2693            \\
BPR (1000) \cite{ICLR16}         & 0.6322               & 0.2467            \\
M4 (1000)                        & 0.6676               & 0.2847            \\
M2 (100)                         & \textbf{0.7129}      & \textbf{0.3091}   \\ \bottomrule
\end{tabular}
\caption{Best performing models in our experiments compared against various baselines.}
\label{rescmp}
\end{table}

\begin{table}[t]
\centering
\begin{tabular}{@{}ccc@{}}
\toprule
\textbf{M (GRU Size)} & \textbf{Prediction time (s)} & \textbf{Parameters} \\ \midrule
M1 - M3 (100)                  & 0.665 ($\pm$ 0.023)                     & 5705384                   \\
M4 (100)                       & 0.366 ($\pm$ 0.022)                     & 1950150                   \\
M1 - M3 (1000)                 & 0.824 ($\pm$ 0.025)                     & 42548684                  \\
M4 (1000)                      & 0.485 ($\pm$ 0.022)                     & 7133250                   \\ \bottomrule
\end{tabular}
\caption{Average batch prediction time in seconds and memory requirements for each proposed model (M) at the prediction phase. Prediction times for M4 includes computing the cosine distance of the predicted embedding against each item's embedding.}
\label{restab}
\end{table}

The performance of each model on the evaluation metrics is summarized in Figure~\ref{resplot}. Overall, M1 and M2 yielded strong performance gains over the reported baseline RNN models. From the results of M1, we also see that training with the entire dataset yields slightly poorer results than training it on more recent fractions of the dataset. This indicates that our recommendation models do need to account for changing user behaviour over time. Our best performing models are reported in Table~\ref{rescmp}. We also list the baseline results reported in~\cite{ICLR16}, including their best RNN based models (\emph{i.e.}, TOP1 and BPR) and two traditional algorithms (\emph{i.e.}, S-POP and Item-KNN). Surprisingly, moving from a GRU of 100 to GRU of 1000 did not significantly improve the performance of our models (M1-M3).

We found that the privileged information model (M3) takes an extremely long time to train; we omitted results for M3 with GRU size 1000 as it could not be trained in reasonable time. We believe the main reason for the drastic increase in training time was the need to (1) compute the soft labels, and (2) compute a corresponding cross-entropy loss against these labels for every mini-batch. This scales very poorly when the number of possible labels is large, as is the case here. Nevertheless, M3 yielded modest performance gains over M1 on the smallest dataset sizes. This is consistent with the use of privileged information in \cite{LUPI}, and suggests that it might be useful in settings where little data is available.

Finally, M4 performs poorly compared to our other models in terms of predictive accuracy (although it still improves over the baseline). We may be able to further improve the accuracy in M4 if better quality embeddings were available as targets. We did not, for example, used any additional information of the items, e.g.~category or brand, that will be available in an online store.

On the other hand, the batch prediction time and model sizes are shown in Table~\ref{restab}. Predictions can be made in M4 using only about 60\% of the prediction time of classification-based models (M1-M3). M4 also has much fewer parameters, and therefore, requires less memory. Together, these are steps towards making RNN models deployable in real recommender systems.

\section{Conclusion}

We have presented, and empirically evaluated, several proposed extensions to a basic RNN model. We showed that it is possible to enhance the performance of recurrent models for session-based recommender systems by using proper data augmentation techniques, and accounting for temporal shifts in user behaviour.
Directions for future work include exploring the tradeoffs of the embedding-based model, and using known features of the items in our the models.
\bibliographystyle{abbrv}

\bibliography{bib}  
%

\end{document}